\documentclass[a4paper,11pt]{article}

\usepackage[T1]{fontenc}
\usepackage{lmodern}
\usepackage[utf8]{inputenc}

\usepackage{geometry}
\usepackage{authblk}
\usepackage{graphicx} 
\usepackage{booktabs,tabularx,caption}

\usepackage{hyperref}

\usepackage{array}
\usepackage{multirow}
\usepackage{ragged2e}
\usepackage{booktabs}
\usepackage{soul}
\usepackage{placeins}
\usepackage{amsmath}

\usepackage[backend=biber, style=numeric, maxbibnames=4, minbibnames=3]{biblatex}
\begin{document}

\title{ptt5-v2: A Closer Look at Continued Pretraining of T5 Models for the Portuguese Language}

\author{Marcos Piau}
\author{Roberto Lotufo}
\author{Rodrigo Nogueira}
\affil{School of Electrical and Computing Engineering,\\
State University of Campinas (UNICAMP)}

\date{}

\maketitle

\begin{abstract}
Despite advancements in Natural Language Processing (NLP) and the growing availability of pretrained models, the English language remains the primary focus of model development. Continued pretraining on language-specific corpora provides a practical solution for adapting models to other languages. However, the impact of different pretraining settings on downstream tasks remains underexplored.
This work introduces $\texttt{ptt5-v2}$, investigating the continued pretraining of T5 models for Portuguese. We first develop a baseline set of settings and pretrain models with sizes up to 3B parameters. Finetuning on three Portuguese downstream tasks (assin2 STS, assin2 RTE, and TweetSentBR) yields SOTA results on the latter two.
We then explore the effects of different pretraining configurations, including pretraining data quality, optimization strategies, and multi-epoch pretraining. Perhaps surprisingly, their impact remains subtle compared to our baseline. We release $\texttt{ptt5-v2}$ pretrained checkpoints and their MonoT5-based finetuned $\texttt{MonoPTT5}$ rerankers on HuggingFace in their respective collections at \url{https://huggingface.co/unicamp-dl}.

\end{abstract}

\section{Introduction}
Transformer-based pretrained language models have established themselves as the core paradigm in the field of Natural Language Processing (NLP). Starting with the advent of BERT \cite{devlin2019bert}, which popularized the ``pretrain, then fine-tune'' approach and the use of the transformer architecture ifself, these models acquire a general-purpose language representation by unsupervised pretraining on extensive corpora of unlabeled text. The dynamics of the pretraining process had been studied in depth by many works, like Raffel et al. \cite{raffel2019exploring}, that introduced T5, and scaled up models to billions of parameters and set new SOTAs across many tasks. The trend towards increasing model sizes and datasets to improve performance motivated studies like Kaplan et al. \cite{kaplan2020scaling} on scaling laws and Hoffmann et al. \cite{hoffmann2022training}, who demonstrated the importance of training data size relative to model size for compute-optimal training regimes; more recently work from Gadre et al. \cite{gadre2024language} specifically examined the influence of extended pretraining on downstream task performance. 

Despite the extensive study of pretraining dynamics, the focus has predominantly been on English, leaving non-English languages less explored. Continued pretraining presents a strategic approach to adapting these models to additional languages and domains using significantly less data and computational resources than training from scratch. This method involves further pretraining on language-specific corpora, which has been shown to substantially enhance model performance on downstream tasks in the target language \cite{Pires_2023, carmo2020ptt5, larcher2023cabrita, campiotti2023debertinha, bertimbau,cui2024efficient}. However, there is a lack of detailed investigations into how different settings during the continued pretraining phase influence on downstream tasks performance, with most studies merely aiming for benchmark-leading results without a thorough examination of the underlying factors.

In this work, we study the continued pretraining of T5 models for the Portuguese language, analyzing the impact of various settings on downstream task performance. Rather than solely focusing on achieving state-of-the-art results, our study also investigates how factors like model size, optimization schedules, and pretraining data quality affect the performance. We continue the pretraning of Google's T5  with up to 3 billion parameters on Portuguese texts. By experimenting with different configurations in the pretraining stage, we observe nuanced effects on downstream tasks, with some settings only marginally outperforming the baselines. Our findings also suggest that while continued pretraining enhances model capabilities, the increments in performance diminish as model size increases.

T5 models \cite{raffel2019exploring} demonstrate adaptability across various natural language processing (NLP) tasks due to their encoder-decoder architecture. This structure enables them to process text for both understanding and generation, providing an advantage over encoder-only models like BERT. While not the focus of this work, T5's adaptability to instruction-based fine-tuning, as seen in FLAN-T5 \cite{chung2022scaling}, also enables effective zero-shot and few-shot applications. These factors, combined with the scarcity of Portuguese pretrained encoder-decoder models, motivate our choice to continue the investigation the T5 architecture in this study.

\section{Related Work}
The T5 model~\cite{raffel2019exploring} is an encoder-decoder transformer, and one of its main innovations was to cast all tasks into a text-to-text format, allowing for a unified approach; scaling up models to 11B parameters, they consolidated the transfer learning approach, setting new SOTAs for GLUE \cite{wang2019glue}, SuperGLUE \cite{wang2020superglue}, CNN/Daily Mail \cite{hermann2015teaching} benchmarks. It was pretrained using the ``span corruption'' objective over the C4 (``Colossal Clean Crawled Corpus'') dataset, where random consecutive spans in the input are replaced by special mask tokens, and the model is trained to predict these corrupted tokens (models with up to 11B parameters). Building upon this foundation, mT5 \cite{xue2021mt5} extended the T5 framework to multilingual settings, having been pretrained on mC4, a multilingual dataset covering 101 languages (models up to 13B parameters). PTT5 \cite{carmo2020ptt5} further adapted T5 for Portuguese by continuing the pretraining of T5 models on the BrWac dataset \cite{wagner2018brwac}. This approach led to significant improvements in downstream Portuguese language tasks, which were further enhanced by a Portuguese tokenizer. For clarity, we'll refer to the work of Carmo et al. as \texttt{ptt5-v1}. Other notable international adaptations of T5/mT5 include it5 \cite{sarti2022it5} (Italian), AfriTeVa \cite{jude-ogundepo-etal-2022-afriteva} (low-resource African languages), AraT5 \cite{nagoudi2022arat5} (Arabic), and plT5 \cite{chrabrowa2022evaluation} (Polish).

Bertimbau \cite{bertimbau}, a popular adaptation of the BERT encoder model, remains influential within Portuguese language modeling. Others exploring encoder architectures include Albertina \cite{albertina}, DeBERTinha \cite{campiotti2023debertinha}, the work of Gomes et al. \cite{gomes2023deep} (which pretrains a Roberta model), and de Morais et al. \cite{de2023sub}. Reflecting a broader trend, numerous recent Portuguese models prioritize decoder-only architectures, such as Sabiá \cite{Pires_2023}, Glória \cite{lopes2024gloria}, Bode \cite{garcia2024introducingbode}, Cabrita \cite{larcher2023cabrita}, and Gervásio \cite{santos2024advancing}. In the encoder-decoder space, the work of Carmo et al. (\texttt{ptt5-v1}) explored adapting T5 models for Portuguese. Beyond generic Portuguese models, several works specialize in custom domains: de Barros et al. \cite{tweetsentbr_sota_emoji} and  BERTabaporu \cite{costa-etal-2023-bertabaporu} were designed for Portuguese social media data, while Bertaú \cite{finardi2021bertau} focuses on financial language.

\section{Methodology}
This section describes the methodology for pretraining and evaluating our key experiments, covering the pretraining dataset, language-specific vocabulary, model architectures, optimization strategies, and finetuning and validation processes for downstream tasks.

\subsection{Unsupervised continued pretraining} \label{main_pretraining_methodology}

As the pretraining data, we utilized the Portuguese segment of the mC4 dataset (hereafter referred to as mC4-pt), comprising approximately 524 GB of uncompressed text across 169 million documents. This dataset is significantly larger than the one used for the pretraining of \texttt{ptt5-v1} models, which originated from the BrWac dataset \cite{wagner2018brwac} and consisted of around 15 GB of text from 7.4 million documents after preprocessing.

We adopted the Portuguese language vocabulary from ptt5-v1. This SentencePiece Unigram tokenizer \cite{kudo2018sentencepiece}, comprising 32,000 tokens, was trained over a corpus of 2 million documents from the Portuguese Wikipedia. This vocabulary shares the same number of tokens and control tokens as T5, facilitating the direct use of Google's model checkpoints.

As the pretraining objective, the span corruption task was employed, utilizing batches of 128 sequences of 512 tokens (65,536 tokens) - a methodology consistent with the baseline experiment by Raffel et al. \cite{raffel2019exploring}. Adafactor optimizer \cite{shazeer2018adafactor} with a constant learning rate of 0.001 and cross-entropy as loss was used during the entire pretraining process. Using these experimental settings, we started from Google's original checkpoints with sizes from \texttt{t5-small} (60M parameters) up to \texttt{t5-3B} (3B parameters), and performed a complete epoch of continued pretraining over the mC4-pt dataset. Considering these settings, a single epoch over the mC4-pt dataset comprises approximately 1,764,515 training steps and 116 billion training tokens. Additional pretraining experiments are detailed in Section \ref{additional_pretrain_experiments_ablation}.

Both pretraining and finetuning experiments utilized TPUv2-8 and TPUv3-8 devices, leveraging \texttt{t5} \cite{raffel2019exploring} and \texttt{seqio} \cite{roberts2022scaling} frameworks.

\subsection{Supervised finetuning on downstream tasks} \label{finetuning_methodology}
We assess the impact of our pretraining on three Portuguese language downstream tasks: ASSIN2 RTE, ASSIN2 STS, and TweetSentBR. The ASSIN2 dataset \cite{ASSIN2} provides two tasks: RTE (Recognizing Textual Entailment), which involves determining whether one sentence entails another, and STS (Semantic Textual Similarity), which quantifies the semantic similarity between sentence pairs on a 1-5 scale. The TweetSentBR dataset \cite{brum-volpe-nunes-2018-building} is a sentiment analysis task for Brazilian Portuguese tweets, classifying them as positive, negative, or neutral. Tables \ref{table:appendix_downstream_datasets} and \ref{table:appendix_downstream_examples} show further details and examples for each task.

\begin{table}[!h]
\resizebox{\textwidth}{!}{%
\centering

\begin{tabular}{lccccc}
\toprule
\textbf{Dataset} & \textbf{Task}            & \textbf{Preferred metric} & \textbf{Train/validation/test} & \textbf{Random score} & \textbf{Possible outputs}
\\
\midrule
ASSIN2 RTE       & Binary classification         & F1-macro                  & 6,500/500/2,448                         &       50                & \{Entailment, None\}
\\
ASSIN2 STS       & Regression                    & Pearson                   & 6,500/500/2,448                         &        0                & $[1,5]$
\\
TweetSentBR      & Multiclass Classification (3) & F1-macro                  & 11,525/1,281/1,982                      &        32.4             & \{negativo, positivo, neutro\}
\\
\bottomrule
\end{tabular}%
}
\caption{Downstream tasks datasets.}
\label{table:appendix_downstream_datasets}

\resizebox{\textwidth}{!}{%
\centering
\begin{tabular}{lcc}
\toprule
\textbf{Dataset} & \textbf{Example inputs} & \textbf{Example targets}
\\
\midrule
ASSIN2 RTE & \texttt{assin2\_rte sentence1: Uma pessoa está escovando um gato sentence2: O pelo de um gato está sendo penteado por uma pessoa} & \texttt{Entailment} \\
ASSIN2 STS & \texttt{assin2\_stsb sentence1: Uma mulher está cortando vegetais sentence2: Uma mulher está cortando brócolis} & \texttt{4.2} \\
TweetSentBR & \texttt{ttsbr\_neg\_pos\_neu\_sentiment\_pt: adorando esse com dr dráuzio varela} & \texttt{positivo} \\
\bottomrule
\end{tabular}%
}
\caption{Downstream tasks inputs and targets}
\label{table:appendix_downstream_examples}
\end{table}

We finetuned the pretrained models over 100 epochs with batches of 128 sequences and a maximum length of 512 tokens, using Adafactor as the optimizer with a constant learning rate of 0.001. The model checkpoint yielding the best performance on the validation set was selected for testing, and greedy decoding was utilized as the decoding method. Because TweetSentBR lacks a validation set, we reserved 10\% of the training data for validation and used the remaining 90\% for training.

All tasks were approached using a text-to-text format. Specifically for the ASSIN2 STS task, which involves the prediction of continuous values in the range between 1 to 5, we adopted the strategy from Raffel et al. \cite{raffel2019exploring}, by rounding the target scores to the nearest 0.2 increment and converting these to strings, thus framing it as a multiclass classification problem compatible with the text-to-text format.

To compare the quality of the new checkpoints against existing alternatives, we also use the same finetuning procedure on Google T5 and mT5 models.

\subsection{MonoPTT5 Rerankers} \label{monoptt5_rerankers}
To evaluate the adaptability of the ptt5-v2 models for information retrieval tasks, specifically passage reranking, we trained MonoT5 rerankers \cite{nogueira-etal-2020-document-monot5} using checkpoints generated as described in Section~\ref{main_pretraining_methodology}. We named these models MonoPTT5. MonoT5 rerankers are used for passage reranking, a two-stage process: first, a less computationally expensive method like BM25 retrieves an initial set of documents for a given query; the reranker model then reranks these documents to improve relevance ordering. The model is trained in a supervised text-to-text manner, learning to generate tokens that correspond to relevant and non-relevant labels. For inference, we greedy decode a single token and calculate the softmax over the logits of the two possible tokens, using the probability of the positive class as the relevance score. We adapted the input and target format for Portuguese to the structure \texttt{"Pergunta: \{query\} Documento: \{document\} Relevante:"}, assigning the token ``Sim'' as relevant and ``Não'' as non-relevant. This format is applied during both training and inference, regardless whether the input language is English or Portuguese.

The training data originated from the mMARCO dataset \cite{bonifacio2022mmarco}, a translated version of the MS MARCO passage retrieval dataset \cite{bajaj2018ms}, originally in English, to 13 languages, including Portuguese. The training subset consists of triples (query, relevant passage, non-relevant passage), which we split into two training example pairs, with each pair containing the query matched to one passage -- either relevant or non-relevant -- thus creating one example for each label. We created a bilingual Portuguese-English training dataset by randomly assigning one of the two languages to each training triplet. This ``translate-train'' approach \cite{conneau2020unsupervised,xue2021mt5,hu2020xtreme} leverages synthetic data augmentation by integrating machine translations with original text data to expand the available training dataset in the target language.  Prior research \cite{bonifacio2022mmarco,rosa2021costbenefit,conneau2020unsupervised} has shown the effectiveness of this bilingual training strategy, motivating our adoption of this method.

The models were trained for 100k steps with batch sizes of 128 sequences and a maximum length of 512 tokens, utilizing Adafactor with a constant learning rate of 0.001 as the optimizer. Instances exceeding the maximum token length were excluded from the training dataset; these constituted approximately 0.01\% of the training samples and were predominantly attributed to noisy translation data. Given the significant computational resources required for training these rerankers, we focused exclusively on models based on the main \texttt{ptt5-v2} checkpoints.

To evaluate the rerankers, we first used BM25 to generate an initial set of relevant documents\footnote{All applications of BM25 in this work use Pyserini's implementation \cite{Lin_etal_SIGIR2021_Pyserini} with default parameters $k_1=0.9$ and $b=0.4$} and then reranked the top-1000 documents. Retrieval metrics are calculated by comparing this ordered list with the relevance judgments from each dataset. We consider two retrieval scenarios: in-domain (using the ``small dev'' set of 6,980 queries from the  mMARCO-pt dataset) and zero-shot (using the Portuguese subset of 249 annotated queries from mRobust\cite{jeronymo2022mrobust04}, and the recently introduced Quati dataset \cite{bueno2024quatibrazilianportugueseinformation}, which has 50 queries). Due to the longer document length in mRobust, we segmented documents into sliding sentence windows using a Spacy \cite{Honnibal_spaCy_Industrial-strength_Natural_2020} sentencizer pipeline, with a maximum length of 8 and a stride of 4 sentences to mitigate truncation during reranking.

We considered the two Quati subsets (1M and 10M documents), both of which share the same 50 queries. Quati differs from mMARCO-pt and mRobust-pt in two key aspects: unlike those translated English datasets, Quati comprises documents originally in Brazilian Portuguese and queries created by native speakers. Additionally, it employs a large language model for relevance judgments on query-document pairs, contrasting with the human annotation methods used in the other collections.

\section{Main Results}
Table \ref{table:main_results} shows the results on the downstream tasks considered. In the ASSIN2 RTE task, our 3B sets a new SOTA, surpassing the current one by 0.61 F1-macro points. For the TweetSentBR dataset, we achieved better performance than current finetuned SOTAs with \texttt{ptt5-v2-large} and \texttt{ptt5-v2-3B}, by 0.52 and 1.54 F1-macro points, respectively, but our results are worse when comparing to GPT-4. We highlight that our \texttt{ptt5-v2} were trained exclusively on each task training data using the text-to-text framework without any data augmentation or adaptation to the model's architecture, unlike the works of \cite{tweetsentbr_sota_emoji} and \cite{rosa2021costbenefit}, which held the SOTA for TweetSentBR and ASSIN2 RTE. In the ASSIN2 STS task, our models did not surpass the current SOTA; regardless, \texttt{ptt5-v2} still shows better performance than mT5 and T5 models with approximate sizes, and this is also the only task where a smaller \texttt{ptt5-v2} model (\texttt{ptt5-v2-large}) has better performance of a large one \texttt{ptt5-v2-3B}.

\begin{table}[htbp]
\resizebox{\textwidth}{!}{%
\centering
\begin{tabular}{lccccccccc}
\toprule
& & Binary classification & Regression & Multiclass classification & \begin{math}\leftarrow \end{math} & \multicolumn{4}{c}{Retrieval} \\ 
\cmidrule(lr){3-3} \cmidrule(lr){4-4} \cmidrule(lr){5-5} \cmidrule(lr){6-6} \cmidrule(lr){7-10} 
& & {ASSIN2 RTE} & {ASSIN2 STS} & {TweetSentBR} & \begin{math}\leftarrow \end{math} & {mMARCO-pt} & {mRobust-pt} & {quati-1M} & {quati-10M}\\
Model & Parameters & F1-macro & Pearson & F1-macro & NPM & MRR@10 & nDCG@20 & nDCG@10 & nDCG@10\\ 
\midrule
\multicolumn{10}{l}{\textit{T5}} \\
\midrule
t5-small & 60M & 83.66 & 0.738 & 62.14 & 61.71 & - & - & - & - \\
t5-base & 220M & 85.80 & 0.764 & 65.43 & 65.63 & - & - & - & - \\
t5-large & 770M & 88.91 & 0.790 & 68.22 & 69.95 & - & - & - & - \\
t5-3B & 3B & 90.78 & 0.827 & 72.58 & 74.56 & - & - & - & - \\ 
\midrule
\multicolumn{10}{l}{\textit{mT5}} \\
\midrule
mt5-small & 300M & 75.36 & 0.688 & 61.81 & 54.35 & - & - & - & - \\
mt5-base & 580M & 79.39 & 0.749 & 70.76 & 63.46 & - & - & 0.5356 & 0.6557 \\
mt5-large & 1.2B & 88.25 & 0.753 & 61.11 & 64.76 & - & - & - & - \\
mt5-xl & 3.7B & 91.81 & 0.827 & 77.05 & 77.45 & - & - & 0.6262 & 0.7051 \\ 
\midrule
\multicolumn{10}{l}{\textit{ptt5-v2}} \\
\midrule
ptt5-v2-small & 60M & 87.14 & 0.782 & 70.99 & 69.86 & 0.273 & 0.344 & 0.4174 & 0.5755 \\
ptt5-v2-base & 220M & 88.36 & 0.814 & 73.20 & 72.82 & \textbf{0.311} & 0.384 & 0.5662 & 0.6756 \\
ptt5-v2-large & 770M & 91.73 & 0.839 & \textbf{76.78} & 77.68 & \textbf{0.315} & \textbf{0.462} & 0.5937 & 0.6904 \\
ptt5-v2-3B & 3B & \textbf{92.68} & 0.829 & \textbf{77.80} & 78.48 & \textbf{0.332} & \textbf{0.512} & \textbf{0.6385} & \textbf{0.733} \\ 
\midrule
Supervised SOTA & - & 92.07 \cite{rosa2021costbenefit} & \textbf{0.868} \cite{albertina} & 76.26 \cite{tweetsentbr_sota_emoji} & - & 0.306 \cite{rosa2021costbenefit} & 0.391 \cite{jeronymo2022mrobust04}  & - & -\\
\midrule
\multicolumn{10}{l}{\textit{Few shot baselines from \cite{Pires_2023} }} \\
\midrule
GPT-4 & - & 90.96 & 0.776 & 82.40 & - & - & -  & - & - \\
GPT-3.5-turbo & - & 88.28 & 0.664 & 74.39 & - & - & -  & - & - \\
\bottomrule
\end{tabular}
}
\caption{Main results after finetuning. For supervised SOTAs, model sizes are as follows: \cite{rosa2021costbenefit} (335M), \cite{albertina} (900M), \cite{tweetsentbr_sota_emoji} (110M), \cite{rosa2021costbenefit}, and \cite{jeronymo2022mrobust04} (580M). NPM results exclude retrieval tasks.}
\label{table:main_results}
\end{table}

In addition to using individual metrics for each task, we also incorporate the Normalized Preferred Metric (NPM) \cite{srivastava2023imitation} to facilitate the evaluation of the overall performance of a pretrained model across multiple tasks. The NPM normalizes a task's preferred metric (e.g., F1-macro for ASSIN2 RTE), assigning a value of 0 to represent random performance and 100 to denote maximum performance. Below is the equation used to calculating the NPM for a given model and set $N$ of tasks:
\begin{equation}
\textup{NPM} = \frac{100}{N} \sum_{i=1}^{N} \frac{\textup{[raw preferred metric]}_i - \textup{[random score]}_i}{\textup{[maximum score]}_i - \textup{[random score]}_i} 
\end{equation}

Given that MonoPTT5 rerankers were exclusively trained starting from \texttt{ptt5-v2} pretrained checkpoints, retrieval tasks were excluded from this evaluation. Therefore, we only considered ASSIN2 RTE, ASSIN2 STS, and TweetSentBR tasks. For each model, we calculate its aggregate performance by first determining the NPM for each task and then computing the average of these values.

Our \texttt{ptt5-v2} models have higher NPM values than T5 and mT5 models with considerably more parameters: for example, \texttt{ptt5-v2-base} is only surpassed by \texttt{t5-3B} ($\sim$13.6x larger) and \texttt{t5-xl} ($\sim$16.81x larger); this performance gap, however, is most pronounced in smaller models, narrowing as model size increases. A similar result was also observed by Xue et al. \cite{xue2021mt5}, which analyzed the performance of T5 and mT5 models peformance on the SQuAD benchmark \cite{rajpurkar2016squad}, observing a peformance gap between \texttt{t5-small} and \texttt{t5-base} vs mT5 models of equivalent sizes, which is diminished starting from \texttt{t5-large}. This performance gap observed on smaller model sizes is advantageous when we consider environments constrained by computational resources, increasing the maximum attainable level of performance; additionally, a language-specific tokenizer reduces text splitting into fewer tokens, leading to lower latency and the potential to accommodate more text within the same maximum token context window. Interestingly, mT5 models tend to show lower NPM values, except in the 3 billion parameter range, where they slightly outperform T5 models.

For the retrieval tasks, our MonoPTT5 rerankers were able to set new SOTAs for both mMARCO-pt and mRobust-pt. For the mMARCO-pt dataset, model sizes starting from \texttt{ptt5-v2-base} were able to surpass the current SOTA; the \texttt{ptt5-v2-3B} reranker obtained a gain of +0.026 points in MRR@10. In the mRobust-pt dataset, our \texttt{ptt5-v2-large} and \texttt{ptt5-v2-3B} rerankers surpassed the current SOTA by +0.071 and +0.121 in nDCG@20, respectively. 

\begin{figure}[ht]
\centering
\includegraphics[width=0.8\textwidth]{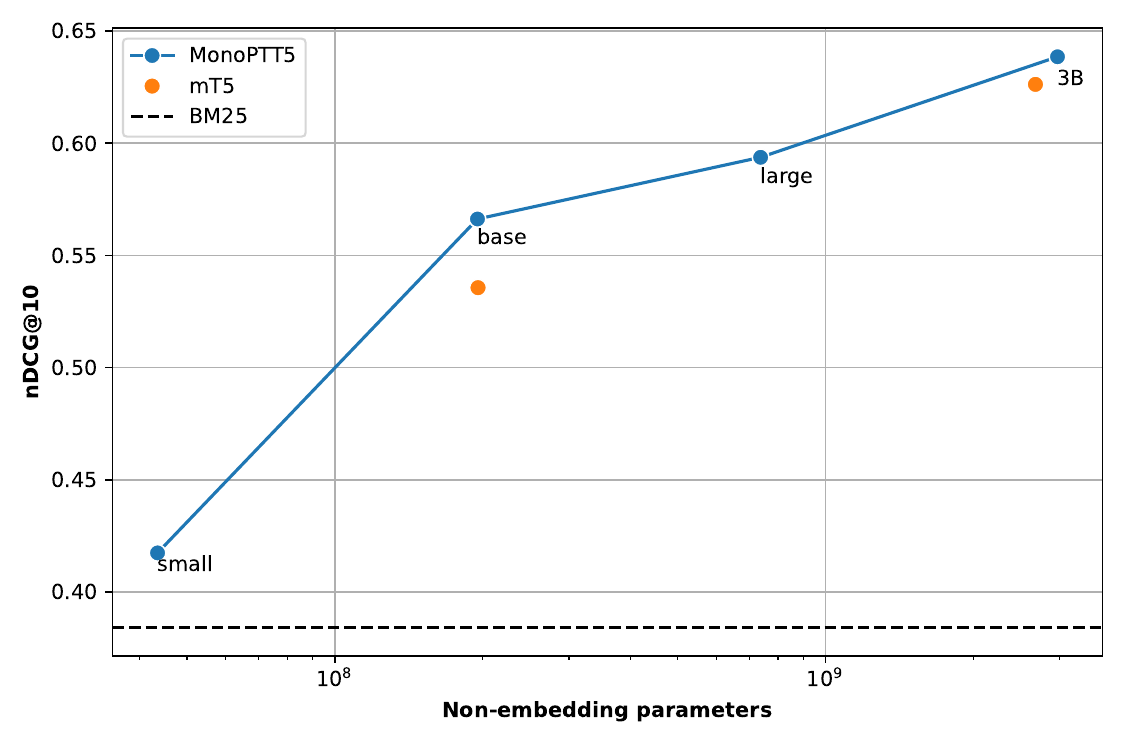}
\caption{Retrieval results on quati-1M. Total size excludes embedding parameters.}
\label{fig:quati_1M}
\end{figure}

\begin{figure}[ht]
\centering
\includegraphics[width=0.8\textwidth]{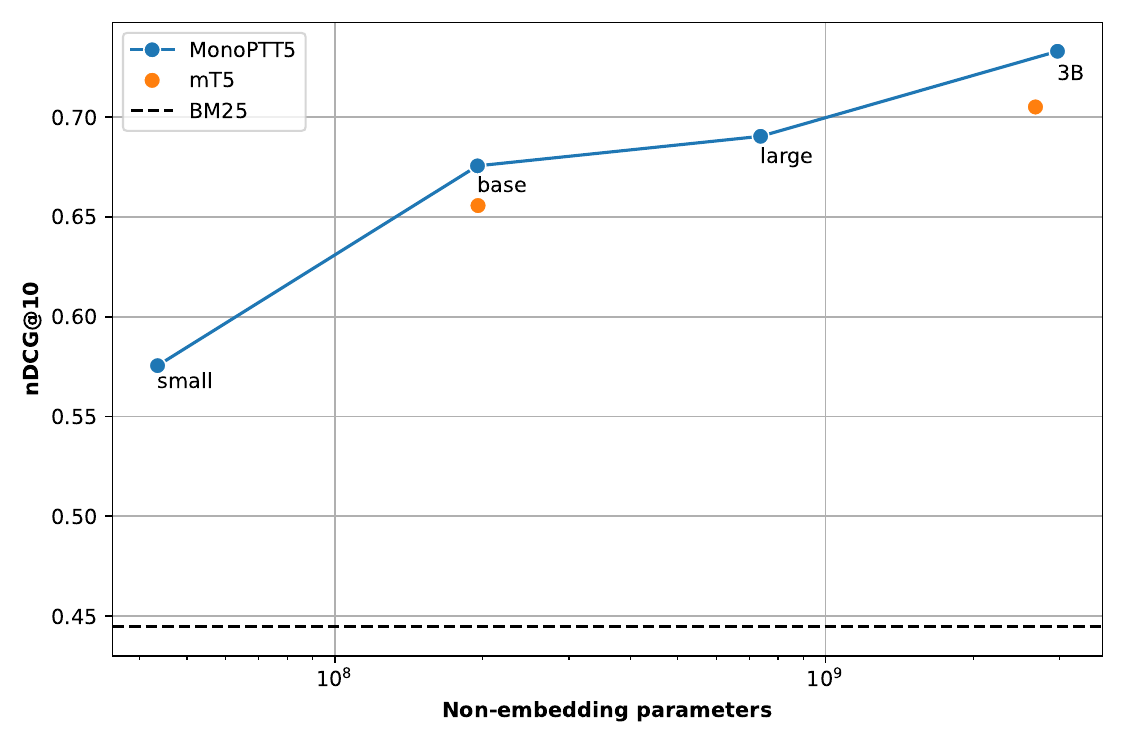}
\caption{Retrieval results on quati-10M. Total size excludes embedding parameters.}
\label{fig:quati_10M}
\end{figure}

Direct SOTA comparisons are not feasible for the Quati dataset since the relevance judgment annotations depend on the complete set of retrieved documents by the compared systems. Nevertheless, our models demonstrated consistent improvements over existing rerankers: the \texttt{ptt5-v2-base} reranker achieved gains of +0.031 and +0.020 points in nDCG@10 compared to the \texttt{mt5-base} reranker from Bonifacio et al.\footnote{\url{https://huggingface.co/unicamp-dl/mt5-base-en-pt-msmarco-v2}} on quati-1M and quati-10M, respectively. Similarly, our \texttt{ptt5-v2-3B} reranker showed improvements of +0.012 and +0.028 points in nDCG@10 over the \texttt{mt5-xl} reranker\footnote{\url{https://huggingface.co/unicamp-dl/mt5-3B-mmarco-en-pt}} for quati-1M and quati-10M, respectively. These comparisons are illustrated in figures \ref{fig:quati_1M} and \ref{fig:quati_10M}. A more detailed analysis of these retrieval tasks is presented in section \ref{additional_rerankers_results_ablation}.

\section{Ablations}
\subsection{Additional pretraining experiments} \label{additional_pretrain_experiments_ablation}
This section includes pretraining experiments additional to those described in \ref{main_pretraining_methodology}.

\subsubsection{Comparison with ptt5-v1}
Given the title of our work, a pertinent question arises: \textit{How do \texttt{ptt5-v2} models compare with our initial work in \texttt{ptt5-v1}}? A few key differences exist in the pretraining of \texttt{ptt5-v1}. Notably, it utilized BrWac, a considerably smaller dataset, and a slightly different pretraining objective (denoising, where some input tokens are masked and the model is trained to predict the original text, rather than span corruption). Additionally, \texttt{ptt5-v1} employed models ranging from \texttt{t5-small} to \texttt{t5-large} with an Adafactor optimizer using a learning rate of 0.003 (threefold larger than our setting). \texttt{ptt5-v1} also explored both T5's original English vocabulary and a Portuguese language-specific tokenizer. In contrast, ptt5-v2 exclusively uses the latter.

\begin{figure}[ht]
\centering
\includegraphics[width=0.8\textwidth]{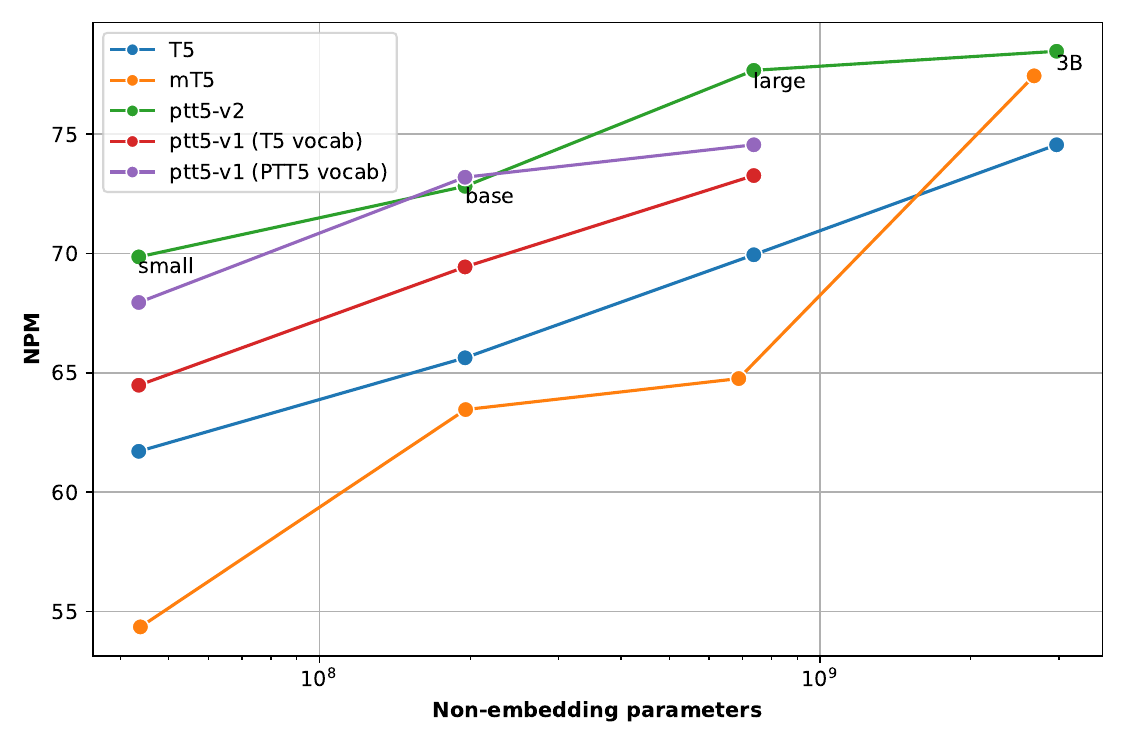}
\caption{NPM vs parameters for varying T5 configurations.}
\label{fig:npm_vs_model_size_main_results_plus_ptt5_v1}
\end{figure}

To ensure a fair comparison, we finetuned the \texttt{ptt5-v1} checkpoints following the methodology outlined in section \ref{finetuning_methodology}. Figure \ref{fig:npm_vs_model_size_main_results_plus_ptt5_v1} presents the NPM values for both \texttt{ptt5-v1} and \texttt{ptt5-v2}, alongside comparisons to mT5 and T5 models. The data corroborates the enhancements achievable through monolingual pretraining on the target language, which is further augmented by employing a dedicated tokenizer. In comparison between the two \texttt{ptt5} iterations, a performance disparity favoring \texttt{ptt5-v2} is apparent for the \texttt{small} and \texttt{large} sizes, with the largest gap observed for \texttt{large}; however, models of the \texttt{base} size exhibit marginally superior performance in the \texttt{ptt5-v1} variant. Surprisingly, mT5's performance lags behind all other models, including the monolingual English T5, except in the 3B parameter range, where it closely matches \texttt{ptt5-v2} and surpasses T5.

\subsubsection{Pretraining data quality}
In our primary experiments, we used the entire mC4-pt dataset for pretraining. This ablation considers two ways to improve pretraining data quality: applying MassiveText \cite{rae2022scaling_gopher} quality filters to mC4-pt and using the BrWac dataset. We restricted this experiment to \texttt{t5-base} models, maintaining the batch size and optimization strategy from the main experiments.

The datasets used in this experiment are built using different approaches:
\begin{enumerate}
    \item \textbf{mC4-pt}: This dataset is derived from the multilingual C4 (mC4) corpus, which is an extension of the original C4 dataset. C4 was built using web data from the Common Crawl\footnote{\url{http://commoncrawl.org}} April 2019 web scrape, while mC4 expanded this by utilizing all 71 monthly scrapes available at the time of its creation. C4 was designed with a focus on English, filtering pages with a minimum probability of 99\% of being English, predicted with \texttt{langdetect}\footnote{\url{https://pypi.org/project/langdetect/}}; mC4 uses \texttt{cld3}\footnote{\url{https://github.com/google/cld3}} for language identification, covering over 100 languages, and keeping only examples with a minimum confidence of 70\%. The C4 dataset pipeline selected only lines finished by English punctuation marks; to ensure correct handling of languages not using these marks, mC4 replaces this step by keeping only pages containing a minimum of three lines with at least 200 characters. The remaining processing is inherited from the C4 dataset: removal of pages containing too few (less than 5) sentences and too short lines (less than 3 words), removal of pages containing obscene words\footnote{\url{https://github.com/LDNOOBW/}}, removal of lines containing the ``javascript'' word, removal of pages containing ``lorem ipsum'' phrase, and pages containing curly brackets. Finally, pages containing the same three-sentence span are considered duplicates and removed. We used only the Portuguese subset of this larger corpus. Notably, this subset is not filtered for specific Portuguese dialects (e.g., Brazilian or European).

    \item \textbf{mC4-pt with MassiveText quality filters}: This version applies Massive Text's filters \cite{rae2022scaling_gopher} to the mC4-pt dataset. Only documents with word counts between 50 and 100,000 words, mean word length between 3 and 10 characters, and symbol-to-word ratio between 3 and 10 are retained. Additionally, documents must have a symbol-to-word ratio below 0.1 (for hashtags or ellipsis), less than 90\% of lines starting with a bullet point, and less than 30\% of lines ending in an ellipsis. Documents with less than 80\% of the words containing alphabetic characters are also removed. We also follow the approach used by Pires et al. \cite{Pires_2023}, removing documents with fewer than two of the translated versions of common stopwords (\textit{the}, \textit{be}, \textit{to}, \textit{of}, \textit{and}, \textit{that}, \textit{have}, \textit{with}); documents containing fewer than 200 unique tokens are also removed. The resulting dataset after applying these filters contains approximately 82 billion tokens, a reduction of about 30\% from the original mC4-pt dataset.

    \item \textbf{BrWac}: This dataset follows the Web as Corpus (WaC) methodology \cite{wacky}, specifically targeting Brazilian Portuguese content. URLs are identified using 8,000 random pairs of medium-frequency Portuguese words from the Linguateca repository \cite{linguateca}, with initial queries to Microsoft Bing's API generating 80,000 seed URLs. These are expanded through two-level link recursion to over 38 million URLs, restricting collection to \texttt{.br} top-level domain pages. Document filtering enforces size constraints (between 5KB and 1MB with minimum length of 256 characters) and requires a minimum of 25\% stopwords to ensure Brazilian Portuguese content, while also removing webpage boilerplate. Duplicate detection uses a modified version of Kilgarriff's \cite{kilgarriff} algorithm, processing documents linearly and removing those with more than 10\% non-original large sentences ($>   $ 25 characters), considering both inter and intra-document duplication. The final step includes linguistic annotation using the Palavras parser \cite{palavras_parser}, though we use only the raw text for our experiments. Beyond BrWac's original pipeline, we normalize text encoding in all documents using \texttt{ftfy} \cite{speer-2019-ftfy}.

\end{enumerate}

Table \ref{table:pretraining_datasets_sizes} summarizes the key characteristics of the resulting pretraining datasets.

\begin{table}[htbp]
    \resizebox{\textwidth}{!}{%
    \centering
    \begin{tabular}{lcccc}
    \toprule
    \textbf{Dataset} &
    \textbf{Training tokens} &
    \textbf{Relative size to mC4-pt} &
    \textbf{Data repetition at one mC4-pt epoch} \\
    \midrule
    mC4-pt &
    115.63B &
    100\% &
    1 \\
    mC4-pt + quality filters &
    82.2B &
    71.1\% &
    1.4 \\
    BrWac &
    3.65B &
    3.2\% &
    32 \\
    \bottomrule
    \end{tabular}
    }
    \caption{Comparison of pretraining datasets: total tokens, data repetition at one mC4-pt epoch, and size relative to mC4-pt.}
    \label{table:pretraining_datasets_sizes}
\end{table}

\begin{figure}[ht]
\centering
\includegraphics[width=0.8\textwidth]{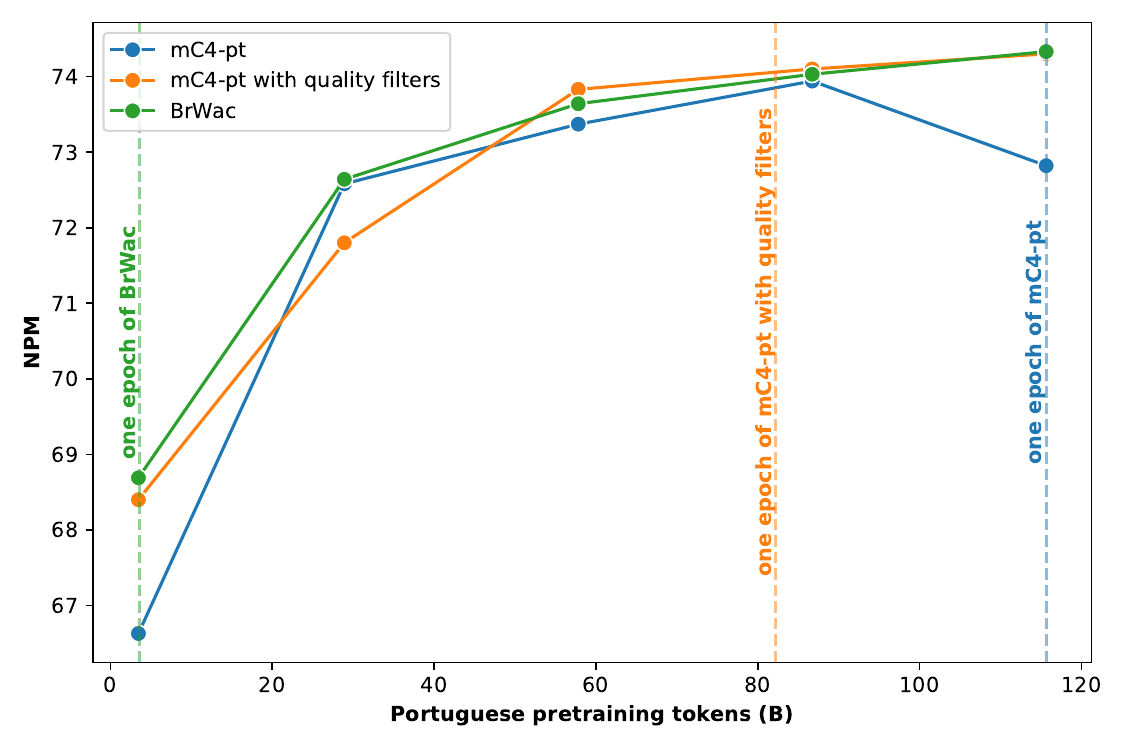}
\caption{Effect of pretraining data quality on downstream task performance.}
\label{fig:mc4_pt_clean_comparison}
\end{figure}

Figure \ref{fig:mc4_pt_clean_comparison} shows the downstream task performance, measured in terms of NPM, for the three different pretraining datasets. Pretraining with the quality-improved datasets demonstrates an upward trend in performance, continuing without saturation up to one mC4-pt epoch -- the last point in the plot. Despite this trend favoring datasets with enhanced quality filtering, the maximum difference in NPM is of only approximately 2 points.

Another interesting observation is that, despite the data repetition for the smaller datasets (approximately 32 times for BrWac), there is no sign of performance degradation. A similar result was observed by Raffel et al. \cite{raffel2019exploring}, who trained models from scratch with varying amounts of data repetition by truncating the pretraining data. They observed limited effects even when data was repeated 64 times, with some benchmarks showing improvements. Their experiments focused on pretraining from scratch, while in our case the repetition occurs during continued pretraining.

\subsubsection{Pretraining optimization strategy}
\begin{figure}[ht]
\centering
\includegraphics[width=0.8\textwidth]{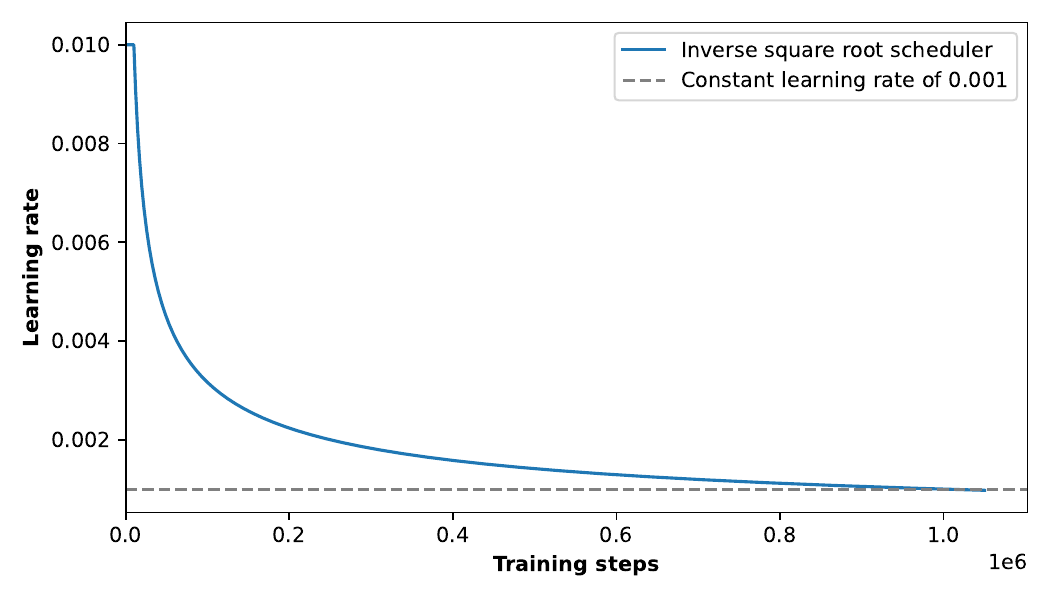}
\caption{Learning rate schedules: Constant vs. inverse square root scheduler as a function of training steps.}
\label{fig:constant_vs_noam_train_steps}
\end{figure}

In our exploration of optimization strategies, we initially employed Adafactor with a constant learning rate. This ablation extends our investigation to the ``inverse square root'' learning rate schedule, as utilized by \cite{raffel2019exploring} in their final pretraining experiments. This learning rate schedule computes the rate as $\frac{1}{\sqrt{\max(n,k)}}$, where $n$ represents the current step, and $k$ is the number of warm-up steps. Raffel et al. \cite{raffel2019exploring} used $k=10,000$, which sets the learning rate of 0.01 for the initial 10k steps, subsequently decreasing exponentially. The learning rate at the end of the pretraining, which consisted of around 1 million steps, was close to 0.001, the same one used during the finetuning. Figure \ref{fig:constant_vs_noam_train_steps} illustrates the difference between these optimization strategies.

\begin{figure}[ht]
\centering
\includegraphics[width=0.8\textwidth]{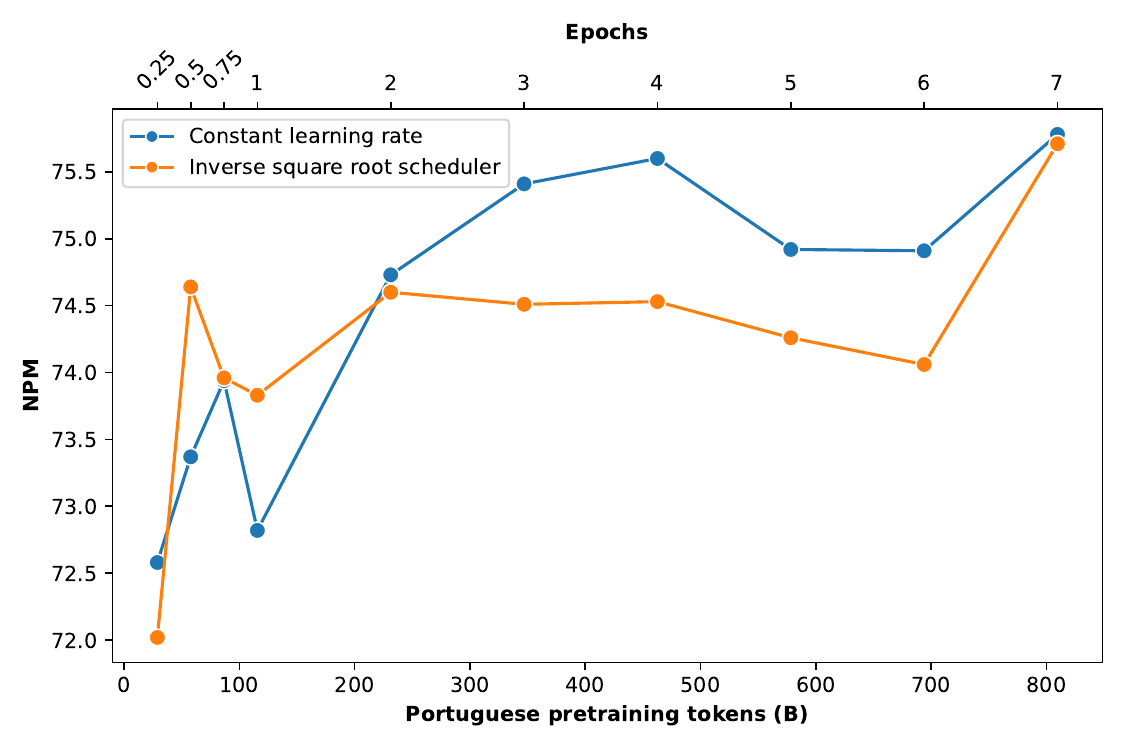}
\caption{Effect of scheduler used during pretraining. Epochs are relative to mC4-pt.}
\label{fig:constant_vs_linear}
\end{figure}

Attempting to closely mirror the T5 pretraining recipe, we applied this identical schedule in our preliminary experiments. However, we observed a rapid overshoot in the training losses for \texttt{t5-large} \texttt{t5-3B} within hours; adjusting $n$ only delayed the overshoot without averting it. Transitioning to a constant learning rate of 0.001 solved the overshoot issue, leading to stable pretraining loss across all model sizes and simplifying our experimental setup. Because overshoot was observed only in the larger models, and also given the computational costs associated with pretraining the larger models, we performed additional pretraining experiments with ``inverse square root'' scheduler for \texttt{t5-small} and \texttt{t5-base} models only.

\subsubsection{Number of pretraining epochs}
In the primary set of experiments, the mC4-pt dataset was fully utilized for pretraining over one epoch. To explore the influence of the number of pretraining epochs on downstream task performance, we conducted experiments over various epochs, including partial epochs (0.25, 0.5, and 0.75 of an epoch). Considering the significant time and computational resources required for extended pretraining, especially with larger models, we limited the pretraining of the \texttt{t5-large} model to 2 epochs and the \texttt{t5-3B} model to a single epoch. The reduced epoch duration for the \texttt{t5-small} and \texttt{t5-base} models enabled more extensive pretraining periods for these configurations.

In Figure \ref{fig:constant_vs_linear}, the NPM values for \texttt{t5-base} are shown with the constant and the inverse square root scheduler, across a varying number of pretraining epochs. It is observed that there is a difference between these two optimization strategies: the inverse square root scheduler has the advantage for up to 2 epochs; afterwards, the constant learning rate takes the upper hand, and by the last epoch considered, they reach the same value. An increasing trend in the values of NPM is also noted for more epochs.

\subsection{MonoPTT5 Rerankers} \label{additional_rerankers_results_ablation}
The information retrieval tasks reported in \ref{table:main_results} represents the performance of our MonoPTT5 experiments, developed with the methodology describe in 
\ref{monoptt5_rerankers}; in this section we also report the results for other approaches, including BM25, and dense retrieval using \texttt{multilingual-e5} \cite{wang2024multilingual} models. The dense models are used as a single stage retrieval system without reranking; dense indexing and retrieval is performed with A100 and V100 GPUs on Google Colab, leveraging the Pyserini framework. For mRobust-pt, which contain longer documents, we mitigate document truncation by using the same splitting strategy described in the section \ref{monoptt5_rerankers}, using the maximum score among the document segments as the document score.

\begin{figure}[ht]
\centering
\includegraphics[width=0.8\textwidth]{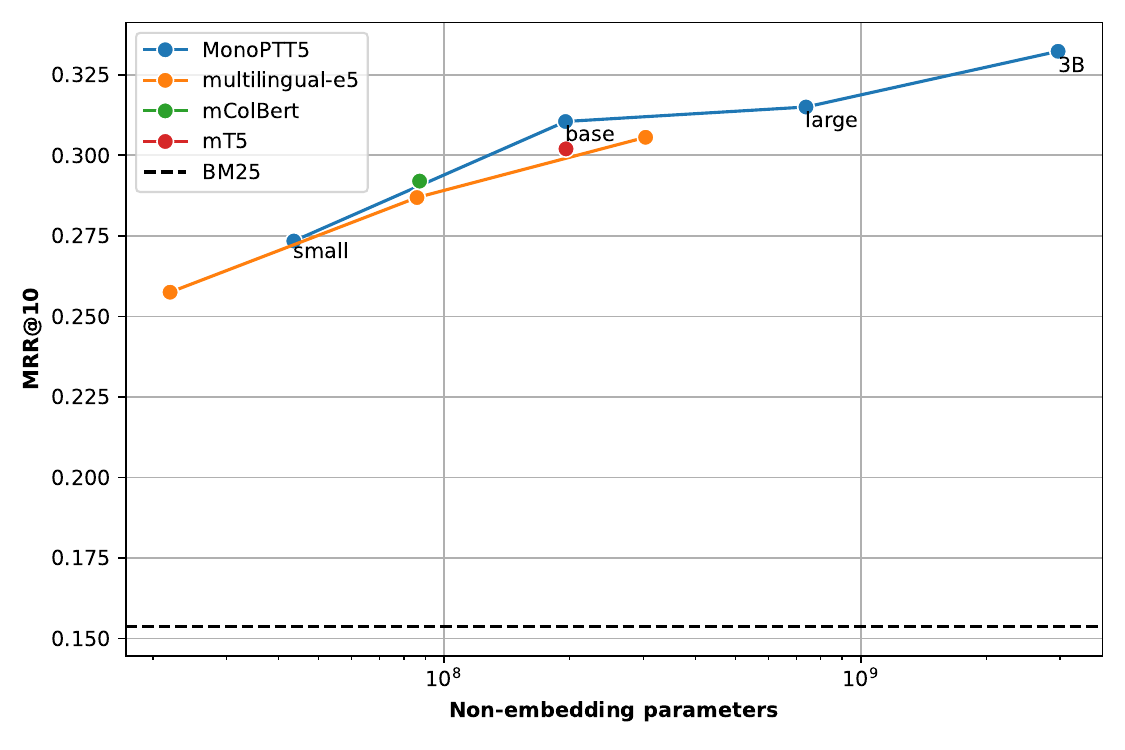}
\caption{Retrieval results on mMARCO-pt. mColbert and mT5 values are from Bonifacio et al. \cite{bonifacio2022mmarco}. Total size excludes embedding parameters.}
\label{fig:mmarco_pt}
\end{figure}

\begin{figure}[ht]
\centering
\includegraphics[width=0.8\textwidth]{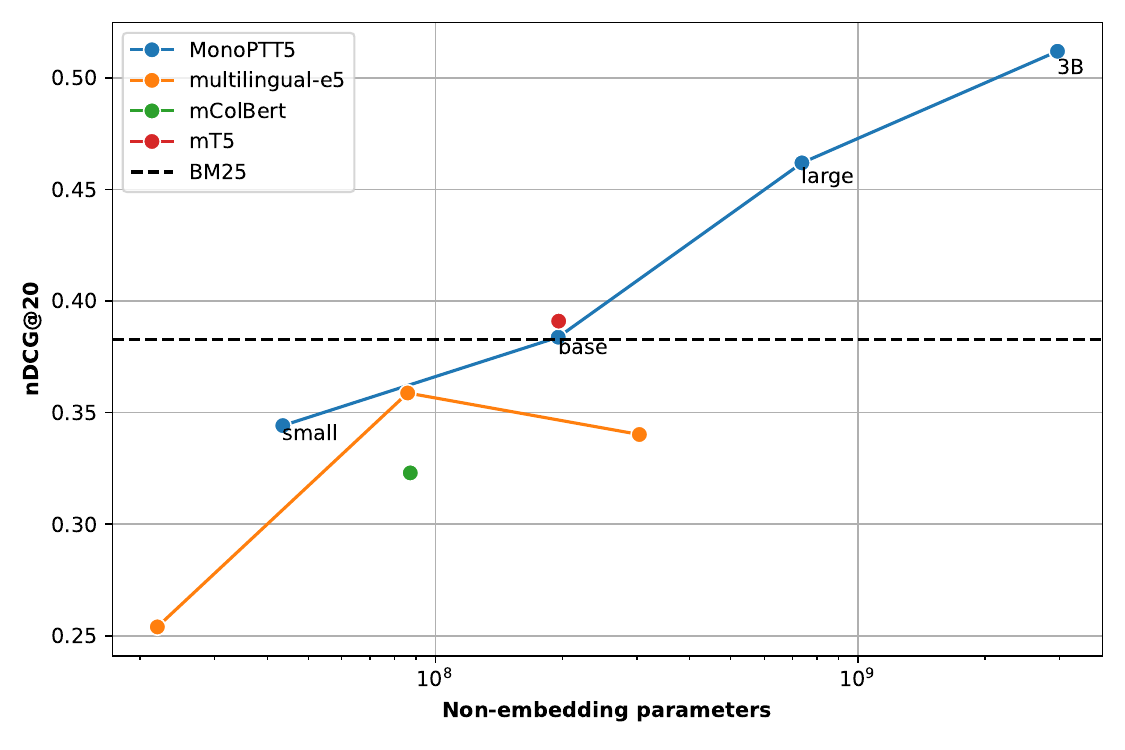}
\caption{Retrieval results on mRobust-pt. mColbert and mT5 values are from Jeronymo et al. \cite{jeronymo2022mrobust04}. Total size excludes embedding parameters.}
\label{fig:mrobust_pt}
\end{figure}

Figures \ref{fig:mmarco_pt} and \ref{fig:mrobust_pt} are used to illustrate the discussion presented in this section. Results in Table \ref{table:main_results} only shows the effectiveness in each retrieval task of our MonoPTT5 models and the SOTA competitors. For the in-domain retrieval task, the mMARCO-pt dataset, we note that BM25 is easily surpassed by all alternatives considered, and the effectiveness figures for MonoPTT5 rerankers and \texttt{multilingual-e5} models are similar when consider the size range in common, and MonoPTT5 rerankers effectiveness is above SOTA starting from models of size \texttt{t5-base}.
For mRobust-pt, representing a zero-shot setting, BM25 is only surpassed by the mT5 reranker from Jeronymo et al. \cite{jeronymo2022mrobust04}, and MonoPTT5 models starting from \texttt{t5-base} size.

\FloatBarrier

\section{Conclusion}
\begin{sloppypar}
In this study, we introduced \texttt{ptt5-v2}, exploring the continued pretraining of T5 models for the Portuguese language. We pretrained T5 models using a Portuguese language tokenizer, over a Portuguese language corpus. The finetuned models achieved SOTA on ASSIN2 RTE and TweetSentBr datasets, two of the three downstream tasks considered. Additionally, we applied these pretrained checkpoints to develop MonoT5 rerankers customized for the Portuguese language, achieving top performance on the mMARCO-pt and mRobust-pt datasets.

Our main results supports the evidence of a performance gap favoring monolingual models over English-focused and multilingual models, a gap that narrows as model capacity increases. This underscores the importance of language-specific pretraining, and our analysis of pretraining settings suggests that while optimization strategies, and pretraining duration can offer incremental improvements, the overall effects were limited in comparison to our baseline settings and the core pretraining recipe remained robust. Regarding pretraining data quality, given the same compute budget, pretraining on a small dataset with high-quality data or a larger dataset with lower-quality data yields models of similar performance on downstream tasks.
\end{sloppypar}

\section*{Acknowledgements}
We thank Google for the TPU grant through the TRC program.

\printbibliography

\end{document}